\documentclass[sn-mathphys]{sn-jnl}

\jyear{2022}%

\theoremstyle{thmstyleone}%
\theoremstyle{thmstyletwo}%

\theoremstyle{thmstylethree}%

\begin{document}

\title[Article Title]{Keeping Minimal Experience to Achieve Efficient Interpretable Policy Distillation}

\author[1]{\fnm{Xiao} \sur{Liu}}\email{liuxiao730@outlook.com}

\author[1]{\fnm{Shuyang} \sur{Liu}}\email{mg20330036@smail.nju.edu.cn}

\author[1]{\fnm{Wenbin} \sur{Li}}\email{liwenbin@nju.edu.cn}

\author[2]{\fnm{Shangdong} \sur{Yang}}\email{sdyang@njupt.edu.cn}

\author*[1]{\fnm{Yang} \sur{Gao}}\email{gaoy@nju.edu.cn}


\affil[1]{\orgname{State Key Laboratory for Novel Software Technology, Nanjing University}}
\affil[2]{\orgname{Nanjing University of Posts and Telecommunications}}

\abstract{
Although deep reinforcement learning has become a universal solution for complex control tasks, its real-world applicability is still limited because lacking security guarantees for policies. To address this problem, we propose \textit{Boundary Characterization via the Minimum Experience Retention (BCMER)}, an end-to-end \textit{Interpretable Policy Distillation (IPD)} framework. Unlike previous IPD approaches, BCMER distinguishes the importance of experiences and keeps a minimal but critical experience pool with almost no loss of policy similarity. Specifically, the proposed BCMER contains two basic steps. Firstly, we propose a novel \textit{multidimensional hyperspheres intersection (MHI)} approach to divide experience points into boundary points and internal points, and reserve the crucial boundary points. Secondly, we develop a nearest-neighbor-based model to generate robust and interpretable decision rules based on the boundary points. Extensive experiments show that the proposed BCMER is able to reduce the amount of experience to $1.4\%\!\sim\!19.1\%$ (when the count of the naive experiences is $10k$) and maintain high IPD performance. In general, the proposed BCMER is more suitable for the experience storage limited regime because it discovers the critical experience and eliminates redundant experience.}

\keywords{Deep reinforcement learning; Interpretable policy distillation}

\maketitle

\section{Introduction}\label{sec-Introduction}
The powerful fitting capability of deep reinforcement learning (DRL) makes it a universal solution for complex problems of learning from interaction~\cite{Sutton2018}, \emph{e.g.}, robot control~\cite{Collins2005}, atari games~\cite{Mnih2013}, and game of go~\cite{Silver2017}. However, the black-box characteristic of DRL will leads to unclear decision-making logic and difficult policy verification~\cite{Bastani2016}, which further results in potential danger on the real-word applications~\cite{Bastani2018}, especially the security-sensitive applications, \emph{e.g.}, air traffic control \cite{Guy2017} and disease diagnosis~\cite{Kao2018}. Therefore, it is urgent to construct a secure, stable, and interpretable guarantee for deep reinforcement learning.

To tackle the above bottleneck, Interpretable Policy Distillation (IPD) policy distillation approaches try to transform the DRL policies into interpretable structures, such as the decision tree~\cite{Bastani2018} and rules~\cite{Lee2019}, using the interactive experiences. Specifically, the IPD approaches normally consists of three basic steps: (1) train a well-performed DRL model as a teacher, (2) collect the state-action pair (or environment transition) and build an experience pool, (3) fit an interpretable model based on the experience pool. By using these explicable structures, the decision-making process is able to be presented in a way that suits humans' thinking. Moreover, by adopting these practical IPD approaches, we also enable the interpretable structure based models to get rewards as high as the DRL models.

\begin{figure*}[htbp]
    \centering\includegraphics[width=\columnwidth]{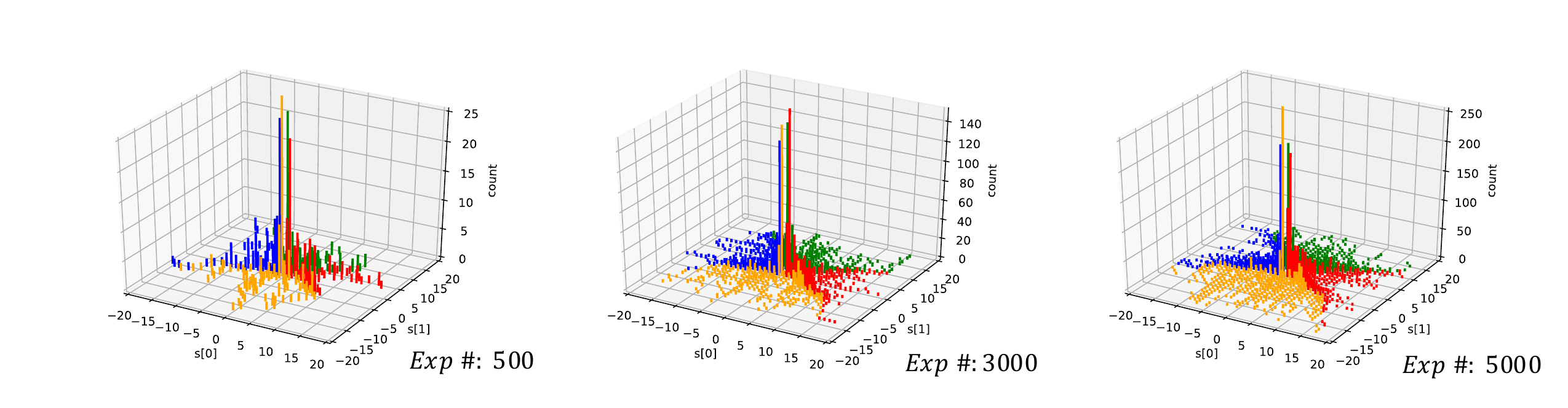}
    \caption{The uneven distribution phenomenon of experience on Predator-Prey. This figure visualizes the interactive experience pool of sizes 500 (a), 3000 (b), and 5000 (c) that were collected from Predator-Prey. The horizontal and vertical coordinates represent the state value, the height represents the accumulated count of the same states, and different colors represent different actions taken by the DRL agent. Thus, most experience distributes around the state $(0,0)$, and collecting experience is challenging for the states away from $(0,0)$.}
    \label{fig:S3-fig1}
\end{figure*}
\begin{figure*}[htbp]
    \centering
    \includegraphics[width=\columnwidth]{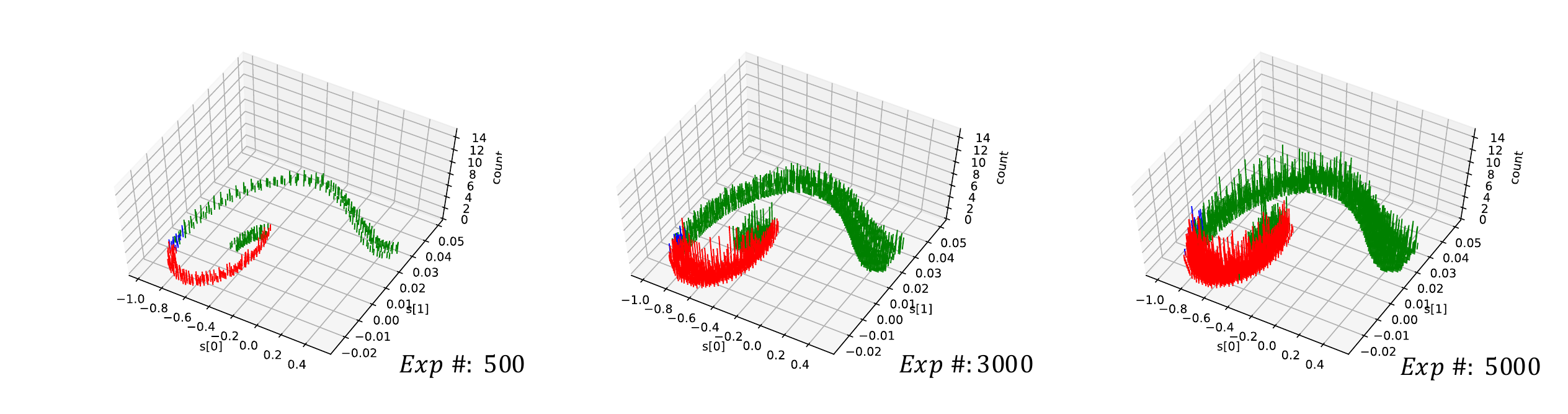}
    \caption{The uneven distribution phenomenon of experience on Mountain Car. This figure visualizes the interactive experience pool of sizes 500 (a), 3000 (b), and 5000 (c) that were collected from Mountain Car. The horizontal and vertical coordinates represent the state value, the height represents the accumulated count of the same states, and different colors are used to distinguish the actions taken by the DRL agent. We cluster states to show the distribution density of states since the status value in Mountain Car is continuous. Experience points accumulate in a ``G-shaped" area. This accumulation of experience did not lead to more knowledge.}
    \label{fig:S3-fig2}
\end{figure*}

It seems that the existing IPD approaches express DRL succinctly. Unfortunately, according to our study of the experience pool, the existing IPD approaches do not consider the characteristic of action preference in DRL models (see section \ref{sec-Obeservation}). Therefore, current IPD approaches generally collect uneven distribution experience points, resulting in a high redundant rate of the experience pool. To be specific, a well-trained DRL model prefers a specific policy. The profound reason for this phenomenon is that, even if the final rewards of two decisions are the same, the Q-value seems to be different. Therefore, DRL policy prefers the action with the highest Q-value when making decisions. For reinforcement learning, it is a property to express the optimal policy through the characteristic of action preference. Nevertheless, for the IPD, the action preference results in highly similar interactive experiences, limiting the knowledge provided by the experience points. In Fig.~\ref{fig:S3-fig1} and Fig.~\ref{fig:S3-fig2}, we visualize the experience pools of Predator-Prey~\cite{Wang2020} and Mountain Car~\cite{Moore1990}. Clearly, in Predator-Prey, most of the experiences are concentrated around $(0,0)$, while in Mountain Car, the experiences concentrate in the G-shaped area. Aside from these two environments that are easy to be visualized, a similar phenomenon occurs in almost all reinforcement learning environments that we know.

Consequently, the action preference phenomenon results in many experiences piled up in the local area of the state space. These piled experience points provide little knowledge related to the model's classification boundaries. In other words, the piled experience points have little contribution to the IPD. On the whole, there are two hidden limitations that the traditional IPD approaches have left:
\begin{itemize}
\item The relationship among DRL's experience points is unclear. Existing IPD approaches generalize rules from data (experience points). For example, when we use CART as the target explicable policy structure, CART generates branches based on information entropy, which reflects the characteristics of the experience population rather than the relationship between experience points. Therefore, it is difficult to say which experience point is more critical and difficult to know what the real DRL's decision boundary looks like. 
\item The cost of experience gathering and storing is not considered. It is usually costly to gain and store experience points in real-world reinforcement learning applications. We cannot store an infinite number of experiences as we do in a simulated environment. Using limited experience points to obtain good distillation performance essentially is a crucial problem that the algorithms need to consider.
\end{itemize}

Different experience points are of different importance. Therefore, in this paper, we divide the experience points into two categories, \emph{i.e.}, boundary points and interior points. Specifically, the boundary points are experience points near the model's decision boundary, containing knowledge of the decision-making boundary. The interior points are experience points inside the decision boundary and are surrounded by the boundary points. For IPD approaches, the boundary points portrayed the DRL decision boundary, thereby contributing a lot to the decision of the target model of policy distillation.

The characteristic of action preference makes many experiences accumulate in local areas, resulting in most experience points becoming interior points. In addition, in most regions of the decision space, the number of boundary points is limited. Therefore, our thinking is to distinguish the two kinds of experience points and retain the boundary points. In this way, we can significantly reduce the experience count while maintain a high decision similarity of distilled policy to the DRL model.

This paper proposes \textit{Boundary Characterization via the Minimum Experience Retention (BCMER)}, an end-to-end IPD framework. Unlike the previous approaches, the proposed BCMER contains an experience selection mechanism, which learns explainable policies using minimal but critical experience points. In other words, the BCMER improves the average quality (or knowledge) of experience points in the case of limited experience pool size. In general, BCMER can be applied in limited experience storage scenarios so that the most critical experience points must be picked up and never exceed the maximum storage load. Because we can not collect and store infinite experience under no circumstances, the BCMER is much closer to the actual applications.

Besides, we design a wide range of experiments to verify the effectiveness of the proposed BCMER. Specifically, we test the experience number, model similarity, and accumulated rewards in four commonly used reinforcement environments, including Predator-Prey~\cite{Wang2020}, Cart Pole~\cite{Barto1983}\cite{Brockman2016}, Mounting Car~\cite{Moore1990}\cite{Brockman2016}, and Flappy Bird~\cite{Urtans2018}. After which, we visualize the experience pools before and after the experience elimination of BCMER. On the whole, in the case of the limited amount of experience storage, the proposed BCMER reduces the experience count to $1.4\%\sim19.1\%$ (when the count of the naive experiences is $10k$). Also, BCMER maintains a high similarity to the teacher model and gets high accumulated rewards.

In summary, our contributions of this work are:
\begin{itemize}
    \item We find that because of the action preference in DRL models, the experience points are unevenly distributed. Therefore, the existing IPD approaches directly collect the DRL's interactive experiences, resulting in a high redundant rate of experience points. 
    \item We propose an end-to-end IPD framework, \emph{i.e.,} \textit{Boundary Characterization via the Minimum Experience Retention (BCMER)}.The BCMER discovers the relative relationship between experience points, constructs the most concise experience pool that characterizes decision boundaries, fits the nearest-neighbor-based model structure that conforms to human logic, and achieves high similarity IPD with minimal storage and interaction costs.
    \item We conduct experiments to verify the effectiveness of the proposed framework. Compared with the baseline approaches, the proposed approach is more practical by remaining high similarity (to the DRL teacher) using fewer interactive experiences.
\end{itemize}

\section{Related Work}\label{sec-Related}

The eXplainable Reinforcement Learning(XRL) is a new application of the eXplainable Artificial Intelligence (XAI). In addition to the difficulties of traditional XAI, the application of XRL is more challenging by confronting the unlabeled characteristics of reinforcement learning. This paper aims to transfer deep reinforcement learning policies to explainable policies based on IPD technology. Therefore, we construct this section with Deep Reinforcement Learning (DRL) and IPD.

\textbf{Deep Reinforcement Learning} \cite{Mnih2013} is a method of expressing and optimizing the agent's policy with a deep neural network that enables the agent to maximize its average accumulated reward in the continuous interaction of the environment. The environmental interaction problems are difficult to use supervised learning since it is hard to get the correct labels for $ S\rightarrow A $. Therefore, traditional DRL problems are modeled based on a Markov Decision Process (MDP), in which the time is divided as separated steps, i.e., time step. Every time step, the agent observe a state $ s_{t}\in S $, and execute an action $ a_t\in A $, Then the state is transformed by the action, i.e., $ s_t\rightarrow s_{t+1} $, and the agent get a reward $ r_t $. Though DRL solves many environment interaction problems well \cite{Collins2005}\cite{Mnih2013}\cite{Silver2017}, people are usually unable to understand the reasons for models to making decisions due to the black box decision-making process of DRLs, so it is hard to judge whether the model is over-fitting. Therefore, a series of explainable studies have been carried out about safe exploration \cite{Moldovan2012}\cite{Wu2016}\cite{Turchetta2016}\cite{Bastani2018}.

\textbf{Interpretable Policy Distillation} \cite{Bastani2018}\cite{Liu2018}\cite{Vasic2019}\cite{Coppens2019}\cite{Dahlin2020}\cite{Loh2011}\cite{Lee2019} is a kind of eXplainable Reinforcement Learning approach that based on the idea of imitation learning \cite{Ross2011}\cite{Hussein2017}. Generally, the idea of IPD is expressing the DRL policies with interpretable model structures. Trees are a kind of commonly used interpretable model structure because of their two characteristics, i.e., nonparametric and highly structured \cite{Bastani2018}. Specifically, its nonparametric makes it the ability to represent complex policies in principle, and it is highly structured, making it interpretable because its hierarchical structure is similar to people's reasoning logic. Though with good characteristics, the trees cannot be optimized like an XRL model because most tree structures do not have gradients. Therefore, imitation learning is introduced to solve the optimizing problem. In general, the existing IPD approaches contains the following three steps: (1) Training teacher model. A well-trained DRL is used as the teacher since it performs well in most environment interaction problems. (2) Collecting experience points. This step collect the interactive experience points of the teacher and the environment, and these experience points are usually stored in two forms, i.e., $ \left\{ (s_1,a_1),...,(s_n,a_n) \right\} $ or environment transition. The experience points are used as the supervised label to build the student model. (3) Fitting student model. Selecting interpretable model structure based on the performance-interpretable tradeoff \cite{Adadi2018}\cite{Tjoa2020}\cite{Puiutta2020} and fitting the interactive experience points. Generally, the higher the upper-performance limit, the less interpretable it is. The distilled policy can replace the XRL model and is one of the few XRL approaches with practicability because of its following two advantages. On the one hand, The goal of IPD is to clone the XRL policy faithfully, so it can be used to verify the knowledge of the original DRL model. On the other hand, the distilled interpretable model usually achieves good performance with a low price, and it can be used as a replacement for DRL.

\section{An Observation: An Uneven Distribution of Interactive Experience}\label{sec-Obeservation}
The existing IPD approaches perform well in austere environments. However, as the environment complexity increases, the interactive experience (or experiences) requirements for these approaches significantly increase, resulting in hindrance in slightly complex environments. Even in Mountain Car \cite{Moore1990}, we need a lot of experience (usually more than $10k$) to get a well-performed interpretable model using current IPD approaches. We believe that only brute increasing the experience pool size is not a good solution. So, we must research the root question, i.e., the relation between the experience pool and the DRL policy. This section analyses the experience distribution characteristics of the existing IPD framework. Specifically, we take Predator-Prey and Mountain Car as examples to demonstrate that collecting experience points with no difference will result in the uneven distribution of experience. 

Reinforcement learning can remember the optimal exploration path. Thus, well-trained reinforcement learning models favor actions with historically higher rewards, i.e., action preference. For DRL models, this ability to remember ensures the stable convergence of reinforcement learning models. However, for approaches that learn from interactive experience, especially IPD, this ability to remember leads to differences between the interactive experience points and DRL policy. To illustrate our point, we visualize the interactive experience of well-trained DRL models in Predator-Prey and Mountain Car environments. As shown in Figure \ref{fig:S3-fig1} and Figure \ref{fig:S3-fig2}, due to the action preference of the DRL model, the interactive experience shows the uneven distribution. Specifically, there are many similar experience points, and these similar experience points are concentrated and distributed in narrow areas of the decision space. This phenomenon occurs not only in Predator-Prey and Mountain Car but also in other environments, and we choose these two environments because they are easy to visualize.

Because of the uneven distribution of experience, when the total number of experiences is constant, this uneven distribution of experience makes some regions have too much knowledge, and others lack knowledge. Therefore, If we collect experience points indiscriminately (like existing IPD approaches), the experience pool will also show an uneven distribution, resulting in a large number of redundant experiences that pile up. This phenomenon of experience's uneven distribution decreases the amount of knowledge that the experience pool provides, leading to the degradation of the performance of the IPD models.
 
Due to the uneven distribution of experience, we believe that the existing IPD approaches have the following problems: (1) Experience Redundant. The existing IPD approaches cannot evaluate the relationship among experience points, resulting in many redundant experience points. (2) Excessive collection of experience. Existing IPD approaches cannot evaluate the relationship between experience and the DRL policy, resulting in excess experience points that must be collected to ensure the integrity of the experience pool.

\section{The Proposed Framework}\label{sec-proposed}
According to the above research, the existing IPD approaches are not perfect because it does not fully consider the action preference characteristics of reinforcement learning. Here comes the basic idea. We believe that experience points near the decision boundaries (boundary points) are more important than others (interior points), where a preliminary proof is shown in section \ref{sec-propose-step2}. Based on this idea, we can avoid the local accumulation of experience by removing the interior points, then fitting the boundary points using a nearest-neighbor-based model, and thus conducting IPD in situations closer to real applications (the experience number is usually limited). Therefore, the critical question is how to efficiently identify the boundary experience points and the interior experience points. Given the above idea, we propose the Boundary Characterization via the Minimum Experience Retention (BCMER), an end-to-end IPD framework. 

\begin{figure*}[htbp] 
    \centering
    \includegraphics[width=\columnwidth]{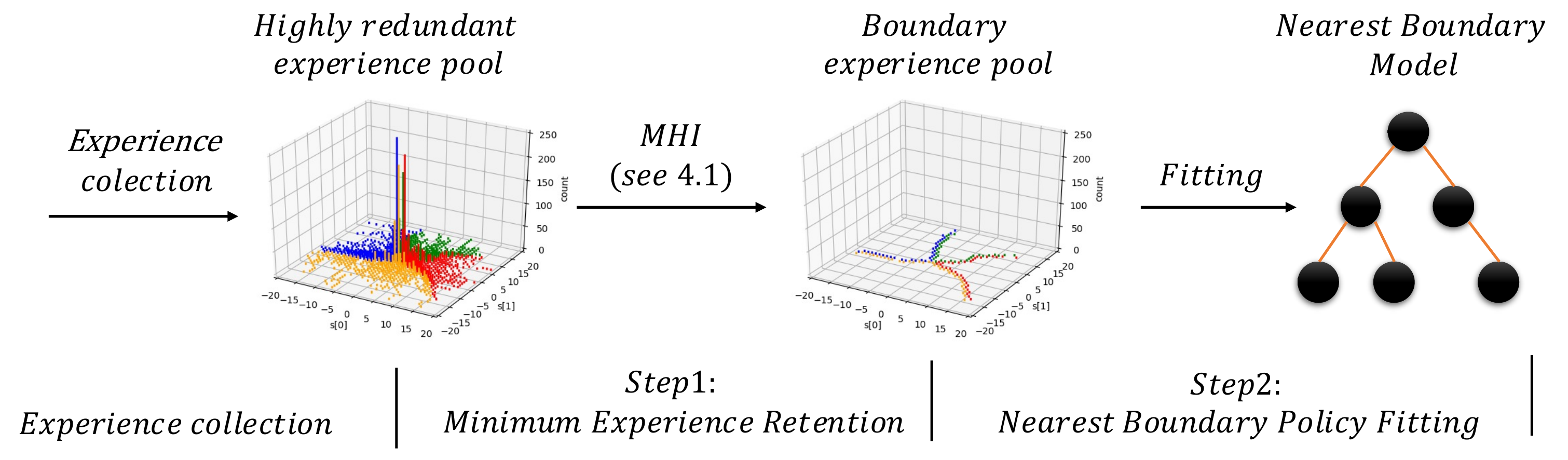}
    \caption{The main workflow of the proposed BCMER framework. The proposed BCMER framework improves on the existing IPD framework. Step 1 simplifies the experience pool to the greatest extent to reduce the stress of experience storage. Step 2 constructs the nearest boundary model (based on the nearest neighbor principle) to realize the explanation of the DRL policy.}
    \label{fig:S4-fig3}
\end{figure*}

As is shown in Figure \ref{fig:S4-fig3}, the proposed BCMER contains two main steps, i.e., minimum experience retention (see \ref{sec-propose-step1}) and nearest boundary policy fitting(see \ref{sec-propose-step2}). In the minimum experience retention step, we identify boundary experience points and interior points and remove internal points to minimize empirical redundancy. Then, in the nearest boundary policy fitting step, we construct a nearest boundary model (based on the nearest neighbor principle) using the boundary experience pool.

\subsection{Step1: Minimum Experience Retention}\label{sec-propose-step1}
We can easily construct a brute solution based on spatial inclusion to distinguish internal points from boundary points. However, this brute solution is challenging to obtain effectively since the computational complexity increases exponentially with the state dimension and experience number.

Specifically, a interior point is surrounded by other points of the same label (action). Assuming that the state has $ 2 $ dimension, and the experience pool is $ E_{o}=\left\{ e_0,e_1,e_2 \right\} $. For any other point $ e $, we can determine whether $ e $ is a interior point by calculate $ f(e) $, i.e.,
\begin{equation}
f(e)=
\begin{cases}
    False. & S_{\triangle e_0 e_1 e_2}<S_{\triangle e e_1 e_2}+S_{\triangle e_0 e e_2}+S_{\triangle e_0 e_1 e} \\
    True. & S_{\triangle e_0 e_1 e_2}=S_{\triangle e e_1 e_2}+S_{\triangle e_0 e e_2}+S_{\triangle e_0 e_1 e}.
\end{cases}
\label{func:S4.1_f1}
\end{equation}
Where $ S_{\triangle e_0 e_1 e_2} $ is the area of the triangle surrounded by $e_0, e_1, e_2$. Although the above method can accurately distinguish interior points, its computational complexity increase with the state dimension and the experience pool size. Specifically, given the state dimension $ n $ and the experience pool size $ C_E $, we must calculate the N-dimensional hyper body for $ N \times C_{C_E}^{n+1} $ times. Therefore, when the number of state dimensions and experience points increases, the method is difficult to obtain a solution in a sufficient time.

\begin{figure}[htbp]
    \centering
    \includegraphics[width=0.7\columnwidth]{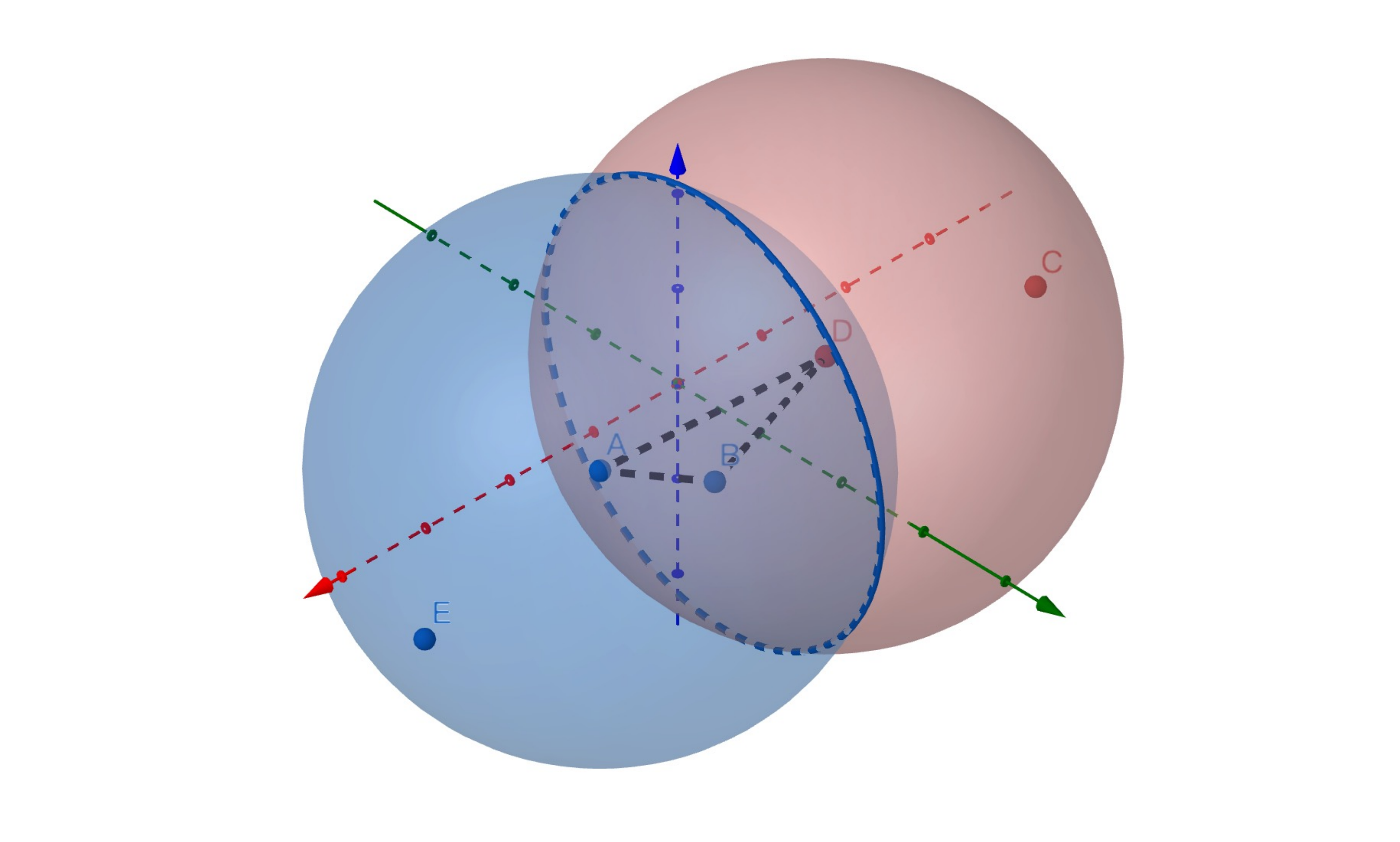}
    \caption{A 3-dimension example of the proposed MHI. In this figure, points $A$, $B$, $E$, and $C$, $D$ belong to different classes. Let us determine if point $A$ is an interior point. Specifically, we first compute the point closest to point $A$ but with a different action, i.e., point $D$. Then, draw the sphere with center point $A$ and $D$, respectively. The sphere radius is equal to the distance between $A$ and $D$. Therefore, the two spheres form an intersection, and when there is $A$ point of the same classification as $A$ in the intersection set, we consider point $A$ as the internal point. In this case, $B$ lies within the intersection of the two spheres, so we identify point $A$ as an interior point.}
    \label{fig:S4-1_fig4}
\end{figure}

To tackle the above bottlenecks, we propose a novel Multidimensional Hyperspheres Intersection (MHI) to judge the boundary point to approximately (see Fig. \ref{fig:S4-1_fig4}). Specifically, for an experience $ e_o\in E $ and the label (action) of $ e_o $ is $ a_{e_o} $. According to Eq. (\ref{func:S4.1_f1}), when a point is contribution point, there must be a point $ e_o'\in E, a_{e_o}\ne a_{e_o'} $that makes experience $ e_o $ and $ e_o' $ be the closest experience with different class (action) to each other. Specifically, for an experience $ e_o\in E $ and the label (action) of $ e_o $ is $ a_{e_o} $. According to Eq. (\ref{func:S4.1_f1}), when a point $ e_o $ is boundary point, there must be a point $ e_o'\in E, a_{e_o}\ne a_{e_o'} $ that makes experience $ e_o $ its closest experience.

Therefore, we propose a three-step approach. First, find the closest point $ e_o' $ of $ e_o $ with $ a_{e_o}\ne a_{e_o'} $, i.e.,

\begin{equation}
e_o'=\{ e_o'\mid  argmin(dis(e_o,e_o')),a_{e_o}\ne a_{e_o'} \}.
\end{equation}
Where $dis()$ is the n-dimension distance function. Then, generate a hypersphere centered on $ e_o' $, and find the experience points contained in the hypersphere, i.e.,
\begin{equation}
\tilde{E}(e_o)=\{ e\mid dis(e,e_o')<dis(e_o,e_o'),a_e=a_{e_o} \}.
\end{equation}
Finally, we determine whether $ e_o $ is a boundary point by checking whether $ \tilde{E} $ is empty, i.e.,
\begin{equation}
f(e_o)=
\begin{cases}
    True.   &   if \quad  \tilde{E}(e_o)=\oslash \\
    False.    &   else.
\end{cases}
\end{equation}

Compared with the brute solution, the proposed MHI avoids the calculation of multi-dimensional hyper-body volume. Since the approach only involves the calculation of distance between experience points, it is easy to implement. So that the MHI can complete calculations in a limited time since the computational complexity does not increase exponentially with the state dimension. The computational complexity of experience screening for the whole experience base is $ O(n^2) $.

\subsection{Step2: Nearest Boundary Policy Fitting}\label{sec-propose-step2}
We believe that IPD succeeds because the decision boundaries of well-trained models have clear rules, and the experience base characterizes these rules. Therefore, it is inefficient only to increase the number of experiences, and the key to improving the performance of the IPD model is to increase the knowledge of the experience pool. This section tries to explain the importance of boundary points and interior points on nearest neighbor-based policy.

Specifically, the boundary points are on the decision boundary of two classes (actions), which draw the range of action boundaries and greatly affect the decision-making. Moreover, the interior points are points far away from the boundaries. In other words, they are in the middle region of a class and are surrounded by boundary points. So that, the boundary points have less impact on decision-making. Giving the state set $ S $ and action sets $ A $. For experience pool $ E $, set $ e_a $ as an experience of class (action) $a\in A$, and $e'$ as the nearest experience of $ e_a $ that labeled $ a'\in A, a'\ne a $, we have
\begin{equation}
\{e\mid dis(s_{e'},s_{e})<dis(s_{e_a},s_{e}),e\in E, a_e=a_{e'}\}==\oslash
\end{equation}
Where $ e^i\subseteq E^i,e^b \subseteq E^b $, $ min() $ is the minimum function. In $ (2) $, when $ \tilde{e}\in E^b $, the label (action) of the nearest experience is $ a_{\tilde{e}^b} $, and when $ \tilde{e}\in E^i $, the label (action) of the nearest experience is $ a_{\tilde{e}^i} $. we have:
\begin{align}
a_{\tilde{e}^i}= a_{e^b}, \quad e^b\in \{ e^b\mid argmin(dis(s,s_{e^b})),e^b\in E^b \} \\
a_{\tilde{e}^b}= a_{e^b}, \quad e^b\in \{ e^b\mid argmin(dis(s,s_{e^b})),e^b\in E^b \}.
\end{align}

Therefore, for any $ s\in S $, regardless of whether the nearest neighbor experience point is boundary point or interior point, the label (action) of the nearest neighbor experience $ a_{\tilde{e}} $ is equal to the label (action) of the nearest boundary point. According to the above discussion, interior points do not affect models based on the nearest neighbor principle. Therefore, the interior points are redundant when using the nearest-neighbor principle imitation learning model. 

In typical applications, too many experience points result in a large model. Therefore, the nearest neighbor model does not have good practical value. However, under the BCMER framework, most experience points (i.e., interior points) are removed. So, the construction of the nearest neighbor model not only avoids the tree being too large but also gives play to the advantage of the quick point finding. We get the suggested actions and the dependent experience points by finding the nearest contribution point. Because all actions are directly based on DRL's historical experience, this approach is easy to verify. After the experience pool ($E$) is divided into the interior and boundary points, we remove all interior points because the interior points will not affect the algorithm based on the nearest neighbor experience points. Then, to speed up the computation, we construct the nearest-neighbor model to fit the boundary points.

\section{Experiments}\label{sec-experiments}
In this section, we design experiments to evaluate the performance of the proposed BCMER framework. Specifically, we evaluated three critical aspects, i.e., policy similarity (see 5.2), experience elimination (see 5.3), and accumulated reward (see 5.4). We believe the policy similarity is the most important property for a IPD approach, and it reflects whether the IPD approach fits the DRL's policy. As for experience elimination, we verify the experience elimination effect of the proposed approach. The uniqueness of the proposed framework is to distinguish and eliminate redundant experiences. Furthermore, the accumulated reward is an essential criterion for RL policies. Because one advantage of IPD is that it can replace the DRL model to interact with the environment, we test the real rewards for our distilled policies in the environment.

\subsection{Experiment Description}\label{sec-experiment-description}
About teacher model: In terms of the teacher model, DQN \cite{Mnih2013} can usually converge to a better policy in the environments where IPD approaches are focusing. Therefore, similar to most IPD approaches, we adopt DQN as the teacher model and learn distilled policy according to the interactive experience between the DQN model and the environment.

\textbf{About baseline approaches(see Table \ref{Table:S5-1_table1}):} The existing IPD framework has yet to consider the selection and optimization of the experience pool. In other words, existing IPD approaches indiscriminate collect interactive experience points. Therefore, the baseline algorithms in this paper do not filter experience. As for the target (student) interpretable model structure, the baseline in this paper takes a widely used Classification And Regression Tree(CART) to fit the experience pool.

\textbf{About proposed approaches(see Table \ref{Table:S5-1_table1}):} The proposed approaches is based on the BCMER framework. Firstly, the collected experience points are divided into boundary and interior points, eliminating all internal points. Then, the nearest boundary model is constructed based on the remaining boundary points, and related methods include Brute-force, KD tree\cite{Sproull1991}, Ball tree\cite{Liu2006}\cite{Omohundro1989}. 
When making a decision, the nearest boundary model searches for the closest boundary point to the current state and returns its corresponding action. Because the nearest-neighbor model is entirely faithful to the experience base, it has steadily improved performance in theory as the number of experiences increases.

\textbf{About environments:} In this paper, we chooses four classic reinforcement learning problems as the experimental environment, i.e., Predator-Prey \cite{Wang2020}, Cart Pole \cite{Barto1983}\cite{Brockman2016}, Mounting Car \cite{Moore1990}\cite{Brockman2016}, Flappy Bird \cite{Urtans2018}. It is worth mentioning that existing IPD approaches are still not suitable for complex environments, so we cover more test environments. In addition, we compile a new Predator-Prey environment with two dimension states to facilitate visual verification. The map size is $ 20 \times 20 $ in our Predator-Prey environment. The agent controls the predator. The predator observes the horizontal and ordinate difference between itself and prey at each step and selects an action according to the observation (i.e., up, down, left, right movement). The prey executes a random policy. At each time step, the prey moves in a random direction, and it has a probability of $ 1/5 $ to stay. At the beginning of each episode, the predator and the prey are born randomly. After each step, if the predator does not catch the prey, it receives a $ -1 $ reward; otherwise, the game is over.

\textbf{About hyper-parameters:} One of the most significant advantages of the proposed BCMER framework is the absence of hyper-parameter. In the main processes, including step 1 (see 4.1) and step 2 (see 4.2), no hyper-parameters must be set. When applying this framework to a new environment, there is no need to introduce prior knowledge.
\begin{table*}[htbp]\small
	\centering
	\caption{The list of models for comparison. Since there is no optimization and selection experience in existing approaches, we choose the commonly used binary-tree structures as the Baseline, and for each structure, shallow (5 layers) and deep (10 layers) structures are implemented. As for the proposed approaches, because they can be applied to all model structures based on the nearest-neighbor principle, we choose three classic structures as examples, and each model structure takes two different layers.}
	\label{Table:S5-1_table1}
	\begin{tabular}{c c c}
	\toprule
		Framework       &Model name 	&Description\\  
	\midrule
	\multirow{1}{*}{/}    
		    &Teacher    &A well-trained DQN.\\
	\midrule
	\multirow{4}{*}{Baseline}   
		    &DT\_Entropy\_l5  &5-layer binary tree that branches by entropy.\\
		    &DT\_Entropy\_l10 &10-layer binary tree that branches by entropy.\\
		    &DT\_Gini\_l5     &5-layer classification and regression tree (CART).\\
		    &DT\_Gini\_l10    &10-layer classification and regression tree (CART).\\
	\midrule
	\multirow{4}{*}{Propose}    
		    &Brute  &Brute-force approach to find the nearest neighbor.\\
		    &KD     &K-Dimensional tree.\\
		    &Ball   &Ball tree.\\
	 \bottomrule
\end{tabular}
\vspace{-3mm}
\end{table*}

\subsection{Policy Similarity}\label{sec-5-2}
IPD realizes policy explanation by migrating policy from a DRL model to an interpretable model. Therefore, faithfully reflecting the DRL's actions is the essential requirement of IPD approaches. In this section, we test Mean Absolute Error (MAE), Root Mean Square Deviation (RMSD), and the decision accuracy (ACC) to evaluate the similarity between the distilled policy. Specifically, the definition of MAE and RMSD is:
\begin{align}
MAE=\frac{\sum_{i=1}^n \mid x_i-y_i\mid }{n}\\
RMSD=\sqrt{\frac{\sum_{i=1}^n (x_i-y_i)^2}{n}}
\end{align}
Where $ x\in X $ and $ y \in Y $ are the output values of the DRL policy and the distilled policy in decision-making, respectively, $n$ is the number of decisions. When the distilled model is close to the original model, it usually has lower MAE and RMSD and higher ACC. Since the state sequence used for testing is generated by the actual interaction of the teacher model and the environment, this experiment considers both DRL's action preference and the actual state transition. Specifically, we first obtain the actual state sequence and DRL's action sequence $ X $ during DRL-Environment interactions. Second, the distilled model generates the distilled action sequence $ Y $ under the actual state sequence.

\begin{figure*}[htbp]
    \centering
    \includegraphics[width=\columnwidth]{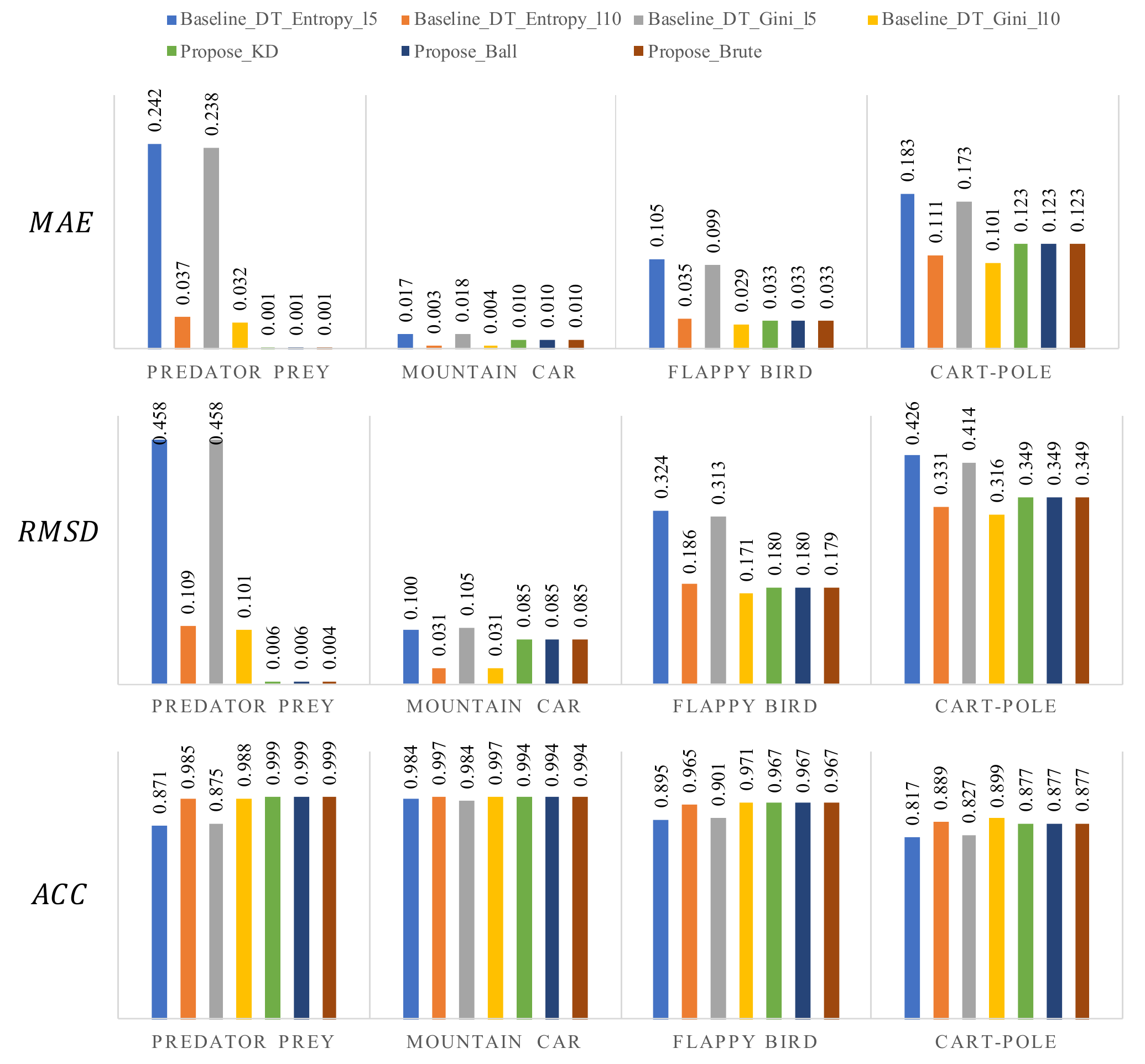}
    \caption{The experiment results of policy similarity. Specifically, the test the policy similarity when the experience number equals 10000. For each experience number, we get its average value of 1000 episodes, and the $ n $ in each episode depends on the actual interaction. The proposed approaches maintain high policy similarity because of the lower MAE and RMSD and the higher ACC. In addition, the policy similarity of the proposed approaches is more stable and without any hyper-parameter.}
    \label{fig:S5-2_fig5}
\end{figure*}

According to Figure \ref{fig:S5-2_fig5}, Even most of the experience is reduced, the proposed approaches still maintain good policy similarities. Specifically, the proposed approaches have the lowest MAE, lowest RMSD, and highest ACC in the environment of Predatory-Prey, which means the proposed approaches are close to the DRL model. Though they do not perform the best in the other three environments, the proposed approaches still maintain a high degree of similarities. Considering that the proposed BCMER removes a significant amount of experience, we believe that the BCMER retains crucial experience points. In addition, the baseline approaches perform unstable similarity in experiments, while the policy similarity of proposed approaches increases steadily with the increase of experience number. Therefore, we believe the nearest neighbor models are more reliable than traditional decision trees since they do not need to fit the data.

\subsection{Experience Elimination}\label{sec-experiment-elimination}
The proposed BCMER preserves only critical experience points (i.e., boundary points), thus theoretically reducing the number of experiences. In this section, we test the empirical reduction of the proposed BCMER. 

\begin{figure*}[htbp]
    \centering
    \includegraphics[width=\columnwidth]{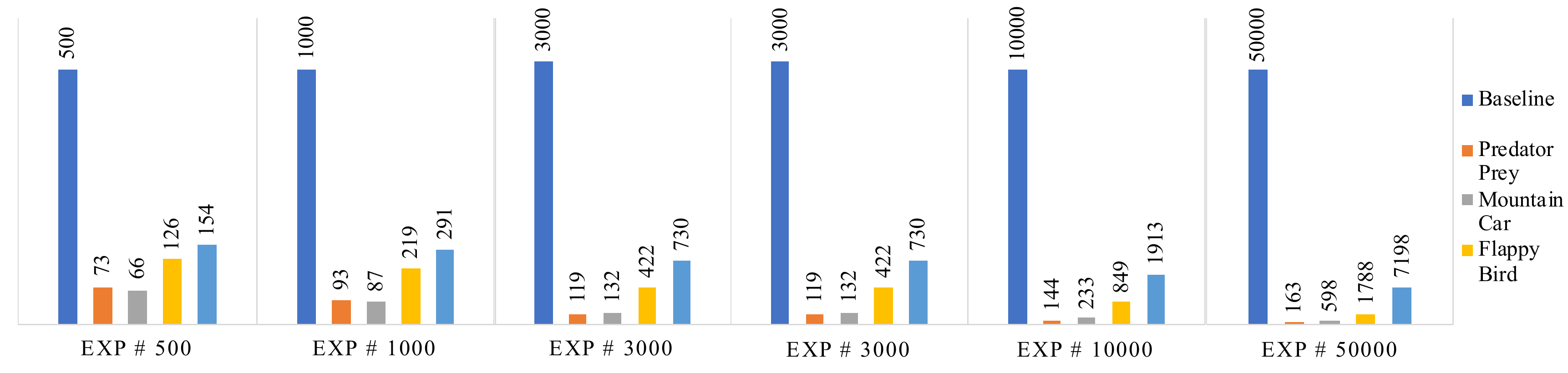}
    \caption{Experiments of experience elimination. This experiment tests experience elimination performance when the experience number is 500, 1000, 3000, 5000, 10000, and 50000. The proposed BCMER can be used directly in different environments and significantly reduce the experience base. And as the total amount of experience increases, the reduction rate increases.}
    \label{fig:S5-3_fig6}
\end{figure*}

Again, we experimented with different environments, and it is worth mentioning that because the proposed BCMER does not have any hyperparameters, it has universal applicability and can be directly applied in different environments. According to the experimental results (Figure \ref{fig:S5-3_fig6}), the proposed BCMER removes most non-critical experience (i.e., interior points). Specifically, the proposed BCMER reduces the amount of experience to $1.4\%$\textasciitilde$19.1\%$ (when naive experience count is $10k$). Combined with Fig. \ref{fig:S6_fig8} and Fig. \ref{fig:S6_fig9}, the proposed approaches preserve experience points near the decision boundary. Even Though most of the experience points are removed, the decision boundaries do not change. Therefore, this experiment can be mutually verified with section \ref{sec-5-2}. In general, the proposed BCMER is more suitable for experience storage limited conditions because it discovers the critical experience and eliminates redundant experience. In addition, critical experience increases slowly as the total number of experiences increases, so the experience reduction rate increases as the total number of experiences increases. This experiment also shows that well-trained DRL models have stable and regular decision boundaries. The decision boundaries of DRL can be delineated using only a few critical experience points.

\begin{figure*}[htbp]
    \centering
    \includegraphics[width=\columnwidth]{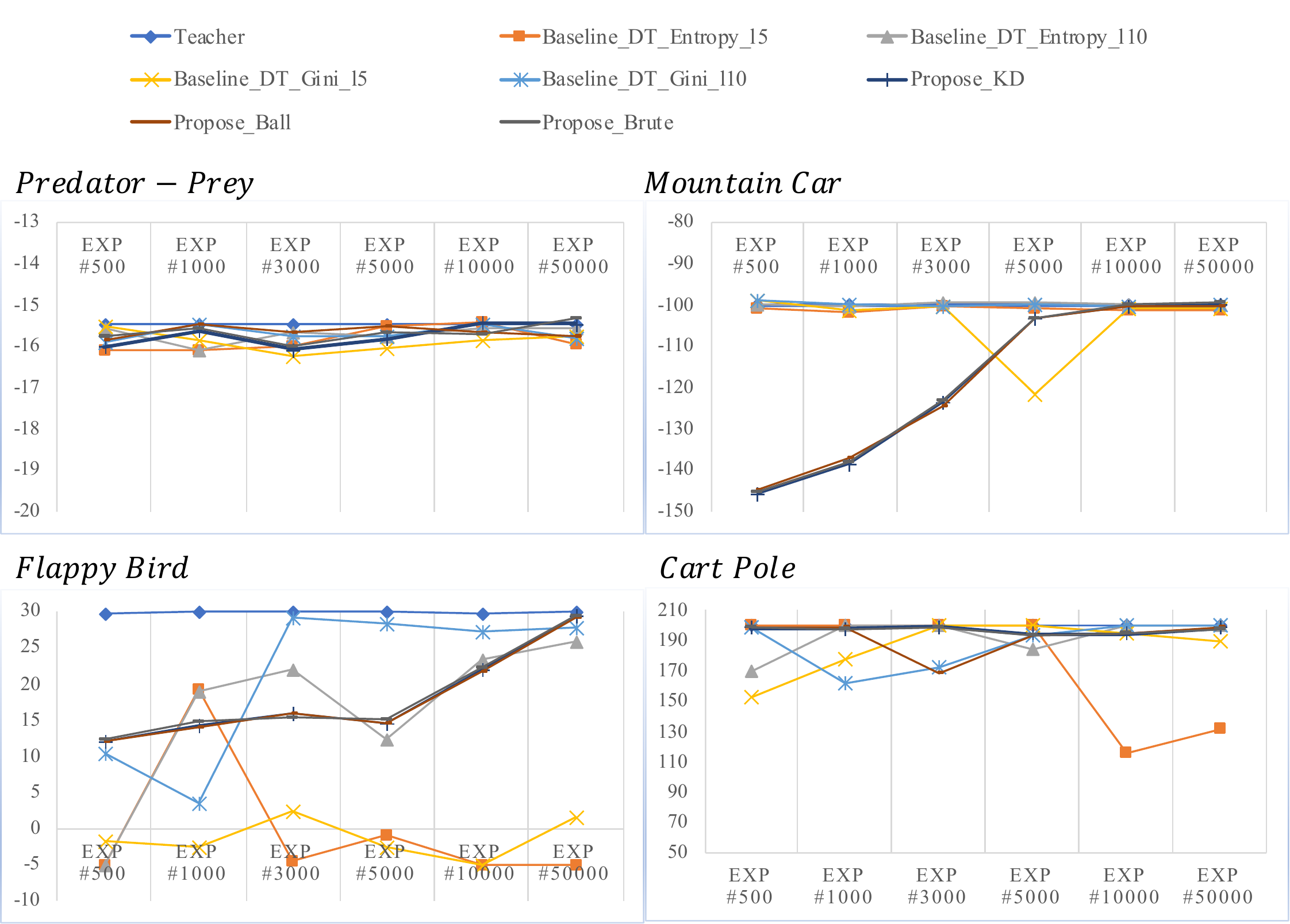}
    \caption{Experiments of accumulated rewards. The experience number is set as 500, 1000, 3000, 5000, 10000, and 50000. Furthermore, each model calculates the average accumulate reward of over 1000 episodes. Although the proposed approaches eliminate most of the experience points, they maintain good performance while being more stable than the baseline approaches that maintain the entire experience pool.}
    \label{fig:S5-4_fig7}
\end{figure*}

\subsection{Accumulated Reward}\label{sec-experiment-reward}
One advantage of IPD is that it can produce decisions. Therefore, testing the performance of the distilled policy in a real environment can measure whether it has learned practical knowledge. In this paper, the proposed BCMER aims to minimize the amount of experience and preserve the performance of the distilled models. This section tests the loss of rewards of the imitated policies.

There is an inevitable performance penalty since the proposed BCMER eliminates most experience points (see \ref{sec-experiment-elimination}). According to the experiment results (see Figure \ref{fig:S5-4_fig7}), we are glad to see that the performance penalty is limited, which means that the proposed framework retains the necessary experience. Specifically, in Predator-Prey, Mountain Car, and Cart-Pole, there was almost no loss of the rewards, whereas, in Flappy Bird, the loss of the reward was acceptable. In addition, compared with the baseline approaches, the proposed approaches have better performance stability and gradually approach the teacher model with the experience increasing. As for the baseline, approaches get unstable rewards, and they may perform poorly when the hyper-parameters are not set very well.

In conclusion, the proposed approaches have limited reward loss and are more stable than the baseline approaches that maintain all experience points. We can generate policies consistent with the teacher model from a few crucial experience points.

\begin{figure*}[htbp]
    \centering
    \includegraphics[width=\columnwidth]{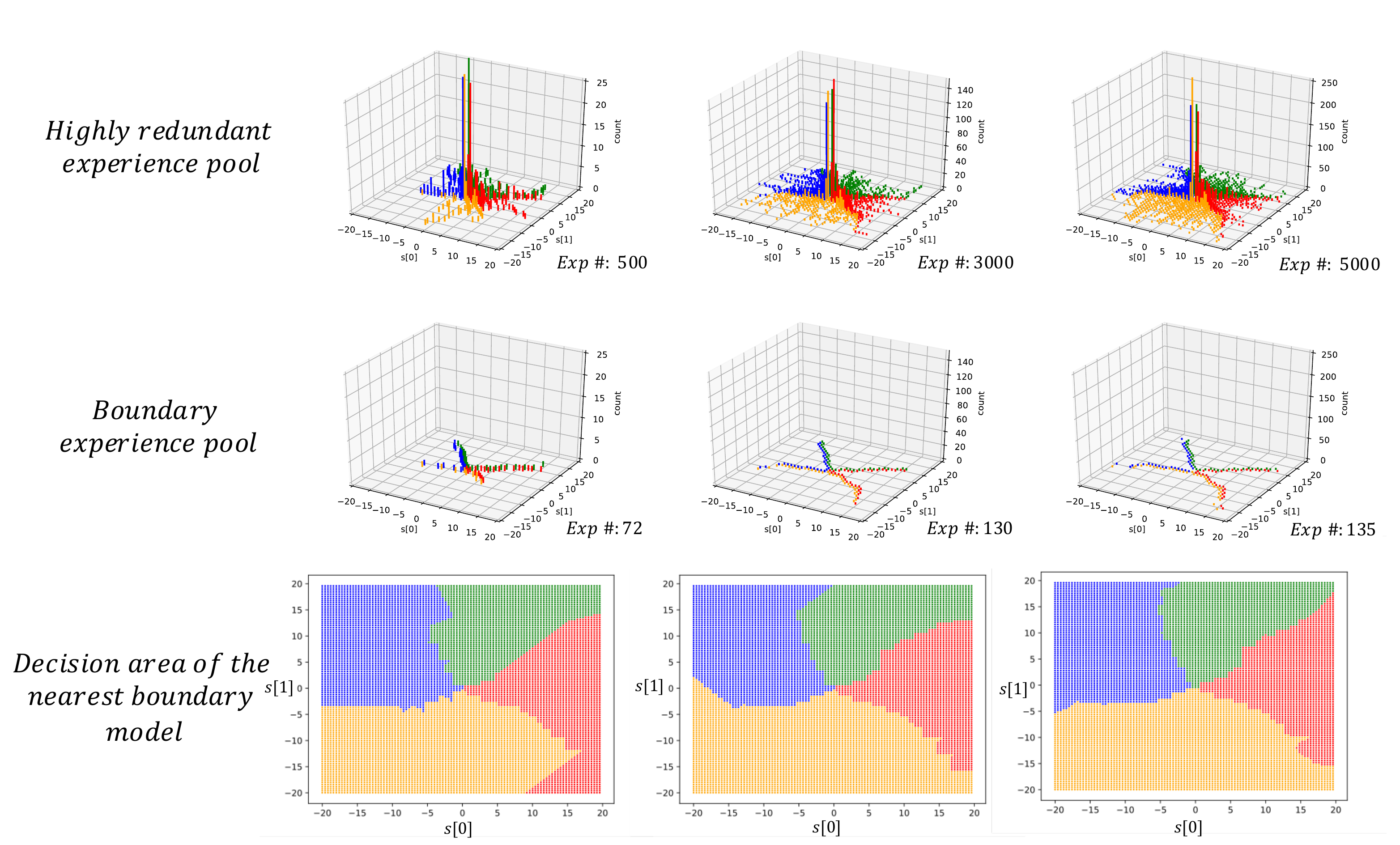}
    \caption{Experience pool visualization Predator-Prey. We visualize the experience pool sizes at 500, 3000, and 5000. There is much redundancy in the naive experience pool, i.e., piled up around state $(0,0)$. As for the experience pool after elimination, most of the experience points are removed, and only the experience points near the decision boundary are retained.}
    \label{fig:S6_fig8}
\end{figure*}

\begin{figure*}[htbp]
    \centering
    \includegraphics[width=\columnwidth]{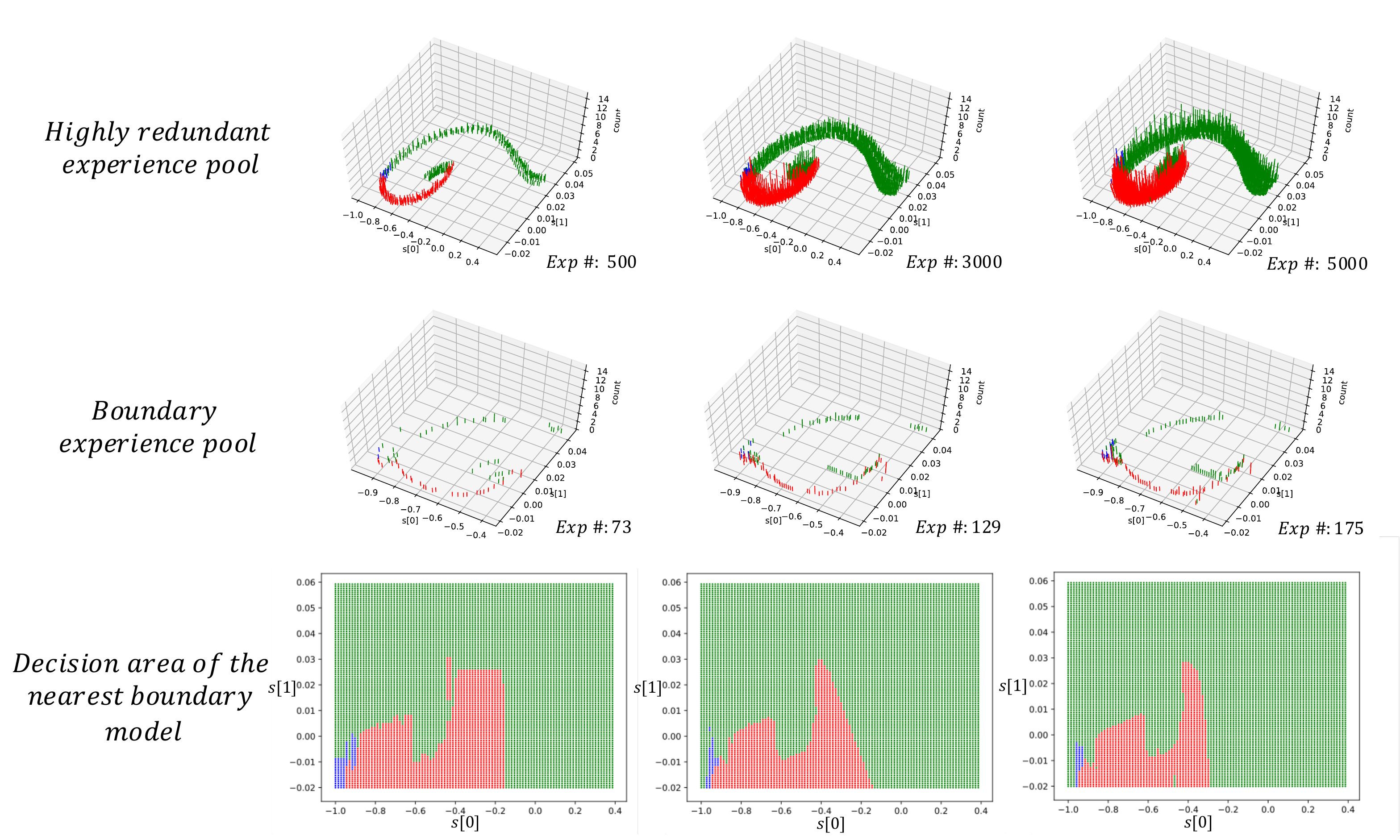}
    \caption{Experience pool visualization of Mountain Car. We visualize the experience pool sizes at 500, 3000, and 5000. Since the states of Mountain Car are continuous, we aggregated similar states and counted their numbers to reflect the density of experience points. After scaling the coordinate axes, we find that experience points present a regular ``G-shaped" distribution. After adopting the proposed BCMER framework, the experience base is greatly reduced, while the decision boundary is not changed.}
    \label{fig:S6_fig9}
\end{figure*}

\section{Visual Verification}\label{sec-verification}
Although we have obtained superior results in the previous experiments, these results are still conceptual. Therefore, we may still have the following doubts, including (1) How does it work? (2) Why does it work? (3) Is it always work? Working in the interpretable domain, we hope to verify the proposed BCMER more intuitively. In this section, we design visual experiments to verify the impact of the proposed BCMER on the experience pool. Specifically, we take Predator-Prey and Mountain Car as examples, visualizing their experience pools before and after the proposed processes. We chose Predator-Prey and Mountain Car because they are easy to visualize (their state dimension is 2), so it is easy to see how the proposed approach affects the experience pools intuitively.

As shown in Figure \ref{fig:S6_fig8} and Figure \ref{fig:S6_fig9}, the naive experience pool is unevenly distributed, and the decision boundary of the XRL model has clear rules. Specifically, in Predator-Prey, experience is concentrated around $(0,0)$, while in Mountain Car, experience is concentrated in a regular ``G-shaped" region. As for the proposed approach, most of the experience points are removed, and the post-processing experience pool maintains the experience points near the decision boundary. Although most of the experiences are removed from the processed experience base, decision boundaries are well preserved. Therefore, the labels (actions) obtained by the nearest neighbor principle-based policy remain unchanged. In addition, we also visualize the decision area of the nearest neighbor boundary model. According to the visualization, with the increase of the number of experiences, the critical experience points (i.e., boundary points) that describe the decision boundaries are increasing, and the nearest neighbor boundary model's decision boundaries become more smooth. This indicates that the knowledge of the nearest neighbor boundary model about the decision boundary is constantly improved, which further makes the policy similarity increase steadily. 

Overall, the visualization of the proposed approach is remarkable, and its results are in line with our expectation of preserving points near the decision boundary. Also, for any interior point that has been eliminated, we can determine its action by looking for its closest boundary point. In addition, with the increase of total experience, the decision boundary is improved rather than revolutionized, proving the proposed approach's stable performance. Therefore, we believe that the proposed approach has simple logic and strong interpretability.

\section{Conclusion}\label{sec-conclusion}
This paper studies the distribution of experience points based on the existing framework of IPD. Because of the action preference of reinforcement learning policy, the experience pools of IPD have a wide phenomenon of uneven distribution. This phenomenon of experience's uneven distribution leads to most empirical distribution in a smaller area, reducing the overall quality of experience. To solve this bottleneck, we aim to study the real influence of experience on the IPD model. This paper proposes an end-to-end Boundary Characterization via the Minimum Experience Retention (BCMER) framework. The BCMER imitates the DRL model using significantly less experience and has high policy similarity and more stable performance compared with traditional approaches. Overall, the significance of this paper is more than the elimination of experience. Furthermore, this paper's work reveals that efficient explainable approaches can be studied via exploring the relationship between experience and decision-making.

\backmatter

\appendix

\begin{thebibliography}{9}

\bibitem{Sutton2018} Sutton R S, Barto A G. Reinforcement learning: An introduction[M]. MIT press, 2018.
\bibitem{Collins2005} Collins S, Ruina A, Tedrake R, et al. Efficient bipedal robots based on passive-dynamic walkers[J]. Science, 2005, 307(5712): 1082-1085.
\bibitem{Mnih2013} Mnih V, Kavukcuoglu K, Silver D, et al. Playing atari with deep reinforcement learning[J]. arXiv preprint arXiv:1312.5602, 2013.

\bibitem{Silver2017} Silver D, Schrittwieser J, Simonyan K, et al. Mastering the game of go without human knowledge[J]. Nature, 2017, 550(7676): 354-359.
\bibitem{Bastani2018} Bastani O, Pu Y, Solar-Lezama A. Verifiable reinforcement learning via policy extraction[C]//Proceedings of the 32nd International Conference on Neural Information Processing Systems (NeurIPS). 2018: 2499-2509.
\bibitem{Guy2017} Katz G, Barrett C, Dill D L, et al. Reluplex: An efficient SMT solver for verifying deep neural networks[C]//International conference on computer aided verification. Springer, Cham, 2017: 97-117.
\bibitem{Kao2018} Kao H C, Tang K F, Chang E. Context-aware symptom checking for disease diagnosis using hierarchical reinforcement learning[C]//Proceedings of the AAAI Conference on Artificial Intelligence. 2018, 32(1).
\bibitem{Bastani2016} Bastani O, Ioannou Y, Lampropoulos L, et al. Measuring neural net robustness with constraints[J]. Advances in neural information processing systems, 2016, 29.
\bibitem{Wang2020} Wang X, Cheng J, Wang L. A reinforcement learning-based predator-prey model[J]. Ecological Complexity, 2020, 42: 100815.
\bibitem{Brockman2016} Brockman G, Cheung V, Pettersson L, et al. Openai gym[J]. arXiv preprint arXiv:1606.01540, 2016.
\bibitem{Urtans2018} Urtans E, Nikitenko A. Survey of deep Q-network variants in PyGame learning environment[C]//Proceedings of the 2018 2nd International Conference on Deep Learning Technologies (ICDLT). 2018: 27-36.
\bibitem{Moldovan2012} Moldovan T M, Abbeel P. Safe exploration in Markov decision processes[C]//Proceedings of the 29th International Coference on International Conference on Machine Learning (ICML). 2012: 1451-1458.
\bibitem{Wu2016} Wu Y, Shariff R, Lattimore T, et al. Conservative bandits[C]//International Conference on Machine Learning (ICML). 2016: 1254-1262.
\bibitem{Turchetta2016} Turchetta M, Berkenkamp F, Krause A. Safe exploration in finite markov decision processes with gaussian processes[J]. Advances in Neural Information Processing Systems (NeurIPS), 2016, 29: 4312-4320.
\bibitem{Liu2018} Liu G, Schulte O, Zhu W, et al. Toward interpretable deep reinforcement learning with linear model u-trees[C]//Joint European Conference on Machine Learning and Knowledge Discovery in Databases (ECML PKDD). Springer, Cham, 2018: 414-429.
\bibitem{Vasic2019} Vasic M, Petrovic A, Wang K, et al. Moët: Interpretable and verifiable reinforcement learning via mixture of expert trees[J]. arXiv preprint arXiv:1906.06717, 2019.
\bibitem{Coppens2019} Coppens Y, Efthymiadis K, Lenaerts T, et al. Distilling deep reinforcement learning policies in soft decision trees[C]//Proceedings of the IJCAI 2019 workshop on explainable artificial intelligence. 2019: 1-6.
\bibitem{Dahlin2020} Dahlin N, Kalagarla K C, Naik N, et al. Designing Interpretable Approximations to Deep Reinforcement Learning with Soft Decision Trees[J]. arXiv preprint arXiv:2010.14785, 2020.
\bibitem{Loh2011} Loh W Y. Classification and regression trees[J]. Wiley interdisciplinary reviews: data mining and knowledge discovery, 2011, 1(1): 14-23.
\bibitem{Lee2019} Lee J H. Complementary reinforcement learning towards explainable agents[J]. arXiv preprint arXiv:1901.00188, 2019.
\bibitem{Ross2011} Ross S, Gordon G, Bagnell D. A reduction of imitation learning and structured prediction to no-regret online learning[C]//Proceedings of the fourteenth international conference on artificial intelligence and statistics (AISTATS), 2011: 627-635.
\bibitem{Hussein2017} Hussein A, Gaber M M, Elyan E, et al. Imitation learning: A survey of learning methods[J]. ACM Computing Surveys (CSUR), 2017, 50(2): 1-35.
\bibitem{Adadi2018} Adadi A, Berrada M. Peeking inside the black-box: a survey on explainable artificial intelligence (XAI)[J]. IEEE access, 2018, 6: 52138-52160.
\bibitem{Tjoa2020} Tjoa E, Guan C. A survey on explainable artificial intelligence (xai): Toward medical xai[J]. IEEE Transactions on Neural Networks and Learning Systems, 2020.
\bibitem{Puiutta2020} Puiutta E, Veith E M S P. Explainable reinforcement learning: A survey[C]//International Cross-Domain Conference for Machine Learning and Knowledge Extraction. Springer, Cham, 2020: 77-95.
\bibitem{Barto1983} Barto A G, Sutton R S, Anderson C W. Neuronlike adaptive elements that can solve difficult learning control problems[J]. IEEE transactions on systems, man, and cybernetics, 1983 (5): 834-846.
\bibitem{Moore1990} Moore A W. Efficient memory-based learning for robot control. PhD. Thesis. Technical Report, Computer Laboratory, University of Cambridge. 1990. No. 209.
\bibitem{Sproull1991} Sproull R F. Refinements to nearest-neighbor searching in k-dimensional trees[J]. Algorithmica, 1991, 6(1): 579-589.
\bibitem{Liu2006} Liu T, Moore A W, Gray A, et al. New Algorithms for Efficient High-Dimensional Nonparametric Classification[J]. Journal of Machine Learning Research (JMLR), 2006, 7(6).
\bibitem{Omohundro1989} Omohundro S M. Five balltree construction algorithms[M]. Berkeley: International Computer Science Institute (ICSI), 1989.

\end{thebibliography}

\end{document}